\documentclass[lettersize,journal]{IEEEtran}
\usepackage{amsmath,amsfonts}
\usepackage{algorithmicx}
\usepackage{algorithm}
\usepackage{array}
\usepackage[caption=false,font=normalsize,labelfont=sf,textfont=sf]{subfig}
\usepackage{textcomp}
\usepackage{stfloats}
\usepackage{url}
\usepackage{verbatim}
\usepackage{booktabs}
\usepackage{graphicx}
\usepackage{cite}
\usepackage{algpseudocode}
\usepackage{multirow} 
\usepackage{color}
\usepackage{todonotes}
\usepackage{color}
\usepackage{transparent}
\usepackage{siunitx}
\usepackage{pifont}
\newcommand{\cmark}{\ding{51}}%
\newcommand{\xmark}{\ding{55}}%

\newcolumntype{P}[1]{>{\centering\arraybackslash}p{#1}}

\newcommand{\Fone}{F1TENTH }

\begin{document}

\title{Unifying F1TENTH Autonomous Racing: \\Survey, Methods and Benchmarks} 

\author{Benjamin David Evans$^{1}$, Raphael Trumpp$^{2}$,~\IEEEmembership{ Graduate Student Member,~IEEE,} Marco Caccamo$^{2}$,~\IEEEmembership{Fellow,~IEEE,} \\Felix Jahncke$^{3}$~\IEEEmembership{Student Member,~IEEE,}, Johannes Betz$^{3}$,~\IEEEmembership{Member,~IEEE,} \\Hendrik Willem Jordaan$^{1}$,~\IEEEmembership{Senior Member,~IEEE,} and Herman Arnold Engelbrecht$^{1}$~\IEEEmembership{Senior Member,~IEEE}

\thanks{$^{1}$ Faculty of Electrical Engineering, Stellenbosch University, Stellenbosch, 7600, South Africa, \texttt{\{bdevans, wjordaan, hebrecht\}@sun.ac.za}.}
\thanks{$^{2}$ TUM School of Engineering and Design, Technical University of Munich, 85748 Garching, Germany, \texttt{\{raphael.trumpp, mcaccamo\}@tum.de}.}%
\thanks{Marco Caccamo is supported by an Alexander von Humboldt Professorship endowed by the German Federal Ministry of Education and Research.}%
\thanks{$^{3}$ The authors are with the Professorship of Autonomous Vehicle Systems, Technical University of Munich, 85748 Garching, Germany; Munich Institute of Robotics and Machine Intelligence (MIRMI)\texttt{\{felix.jahncke, johannes.betz\}@tum.de}.}%
\thanks{Manuscript received \today}}%


\maketitle

\begin{abstract}
The F1TENTH autonomous driving platform, consisting of 1:10-scale remote-controlled cars, has evolved into a well-established education and research platform.
The many publications and real-world competitions span many domains, from classical path planning to novel learning-based algorithms.
Consequently, the field is wide and disjointed, hindering direct comparison of developed methods and making it difficult to assess the state-of-the-art.
Therefore, we aim to unify the field by surveying current approaches, describing common methods, and providing benchmark results to facilitate clear comparisons and establish a baseline for future work.
This research aims to survey past and current work with F1TENTH vehicles in the classical and learning categories and explain the different solution approaches.
We describe particle filter localisation, trajectory optimisation and tracking, model predictive contouring control, follow-the-gap, and end-to-end reinforcement learning.
We provide an open-source evaluation of benchmark methods and investigate overlooked factors of control frequency and localisation accuracy for classical methods as well as reward signal and training map for learning methods.
The evaluation shows that the optimisation and tracking method achieves the fastest lap times, followed by the online planning approach.
Finally, our work identifies and outlines the relevant research aspects to help motivate future work in the F1TENTH domain.
\end{abstract}


\section{Introduction}

The growing field of F1TENTH autonomous driving offers an exciting testbed for cutting-edge robotics and autonomous vehicle research \cite{OKelly2019F1TenthPlatform}.
The domain encompasses diverse aspects of autonomous systems, including perception, planning, control, and learning, and especially focuses on adversarial and high-speed environments \cite{Betz2022AutonomousRacing}.
Autonomous racing presents a uniquely demanding testbed for advancing autonomous robotics methods.
The racing problem is to select control actions (speed and steering) based on raw sensor data (LiDAR) that move the vehicle around the track as quickly as possible.
The nature of racing provides a difficult challenge due to non-linear tyre dynamics, unstructured sensor data and real-time computing requirements \cite{Betz2022TUMChallenge}.
A key challenge is the fundamental conflict between maximising performance and ensuring safety \cite{Liniger2015OptimizationCars}.
Higher speeds increase operational risk while prioritizing safety, which sacrifices competitive advantage.
These difficulties promote the improvement of high-performance, safe, and efficient autonomous vehicle algorithms.

\begin{figure}[t]
    \centering
    \includegraphics[width=0.95\linewidth]{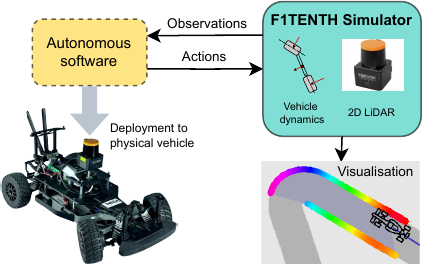}
    \caption{The F1TENTH platform provides a development platform for autonomous algorithms with an accurate simulator allowing rapid deployment.}
    \vspace{-3mm}
    \label{fig:intro_problem_despription}
\end{figure}

The \Fone platform is ideal for research because it enables rapid prototyping of autonomy algorithms, provides a competitive environment and simplifies software-to-hardware deployment \cite{Okelly2020F1TENTH_evaluation}.
F1TENTH vehicles, shown in Fig. \ref{fig:intro_problem_despription}, are equipped with a 2D-LiDAR for sensing, an onboard computer for processing, and motor controllers \cite{OKelly2019F1TenthPlatform}.
The smaller size of F1TENTH cars makes them cost-effective and safe for developing new high-performance algorithms where crashes are inevitable.
Finally, it provides a fun topic for students worldwide \cite{betz2022teaching}, and competitions encourage novel research.

However, the breath of F1TENTH research has left the field divided and fragmented, with researchers often focusing on specific subproblems in isolation.
A particular divide exists between classical and machine learning approaches.
This fragmentation hinders comparison between methods, making it difficult to assess algorithmic advancement.
This paper aims to unify F1TENTH research by providing a comprehensive overview of the field and introducing a consistent perspective to unite different disciplines.
We achieve this by creating a common solution taxonomy through the following contributions:
\begin{itemize}
    \item We provide a literature survey of advances in classical and learning-based approaches that leverage the F1TENTH vehicle. 
    \item We describe common classical and reinforcement learning baseline methods 
    \item We provide benchmark results with open-source software\footnote{Code available at: \url{https://github.com/BDEvan5/f1tenth_benchmarks}} to accelerate future work.
    \item We describe promising research directions for future work on F1TENTH cars to improve current methods and increase the applicability of approaches.
\end{itemize}


\section{Literature Survey} \label{sec:literature_survey}

\subsection{Classical Approaches} \label{subsec:lit_classic_racing}

\begin{figure}[t]
    \centering
    \includegraphics[width=\linewidth]{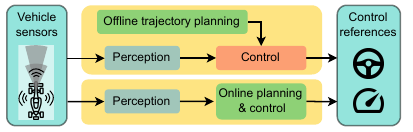}
    \caption{Classical approaches use either offline planning with a separate control module, or online planning and control.}
    \label{fig:racing_pipeline}
\end{figure}

Classical approaches to autonomous racing use a sense-plan-act structure, where the problem is split into distinct sub-problems.
Typically, estimation is used to localise the vehicle on the map, optimisation is used for planning, and tracking algorithms are used to follow the trajectory.
Fig. \ref{fig:racing_pipeline} shows how classical pipelines use localisation with either offline planning and online control or online planning and control.
We study works related to each of these functions.

\textbf{Perception:}
Perception is the task of building a map, localising the vehicle on the map, and perceiving objects.
While no methods of map building, i.e., simultaneous localisation and mapping (SLAM), have been built explicitly for F1TENTH racing, the ROS SLAM toolbox (using graph SLAM) is commonly used \cite{evans2023comparing, gupta2022optimizing}.
Walsh et al. \cite{WalshCDDT:Localization} implement a computationally efficient method for particle filter localisation.
Their approach, and open-source repo, is the standard localisation method used in many other approaches \cite{o2020tunercar, SukhilAdaptiveRacing}.
An extension of their approach has improved the original implementation by using a higher-order motion model \cite{lim2024robustness}.
A different approach using invertible neural networks to learn a localisation policy boasted improved computational performance \cite{zang2023local_inn}.
However, the particle filter remains the most commonly used method due to its simplicity and robustness.

\textbf{Offline trajectory planning (global planning):}
Offline trajectory planning uses an optimisation algorithm and map of the track to generate an optimal set of waypoints.
The baseline planning method for most works uses a shortest path, minimum curvature or minimum time trajectory planning formulation from Heilmeier et al.~\cite{Heilmeier2020MinimumCar}, which formulates the trajectory planning problem as a quadratic programming problem.
Other approaches have turned to genetic programming algorithms to jointly optimise the trajectory and the controller \cite{o2020tunercar, klapalek2021car}, but these can lead to simulator exploitation and unrealistic behaviour \cite{zheng2022gradient}.
Heilmeier et al.'s method is the most popular because it is efficient to compute, robust to different tracks, open-source accessible and provides high-quality results \cite{klapalek2021comparison, becker2023model}.

\textbf{Online trajectory planning (local planning):}
At each planning step, online trajectory planning approaches generate a set of control inputs and a corresponding trajectory for a receding horizon \cite{curinga2017autonomous}.
Model predictive contouring control (MPCC) has been used to optimise a trajectory that maximises progress along the centre line \cite{bhargav2021track}.
Since online planning algorithms replan at each timestep, they can more easily integrate obstacle avoidance and head-to-head racing strategies \cite{bhargav2021track, li2022real}.
While online planning algorithms, such as MPCC, can plan accurate, high-performance paths, they are limited by requiring accurate vehicle models and system identification \cite{ghignone2023tc}.
Additionally, due to the complexity of implementing optimisation routines, there is little comparison between Model Predictive Control (MPC) formulations (such as between different dynamics models) or other methods.

\textbf{Control:}
The controller generates actuator commands for speed and steering that cause the vehicle to track the reference trajectory.
The most common control algorithm is pure pursuit, a geometric path tracking algorithm based on the vehicle dynamics \cite{coulter1992implementation}.
The method has been advanced by using a speed-dependant lookahead distance \cite{SukhilAdaptiveRacing}, and integrating the tyre dynamics into the algorithm to improve tracking at high speeds \cite{becker2023model}.
While MPC is often used as a local planning method, it can also be used as a control approach \cite{nagy2023ensemble, cataffo2022nonlinear}. 
When using MPC for control, the objective is to track the global plan as accurately as possible.
Using MPC for control has the advantage that the system identification can be performed online \cite{wang2021deep}, which means that as the model accuracy can be continuously improved as more data is collected \cite{Jain2020BayesRace}.
Despite these advantages, the pure pursuit algorithm remains the most common control algorithm due to its robustness and simplicity \cite{o2020tunercar, Okelly2020F1TENTH_evaluation, dwivedi2022continuous}.

\begin{table*}[!ht]
    \centering
    \scriptsize
    \renewcommand{\arraystretch}{1.4}
    \caption{Comparative analysis of autonomous racing methods, evaluation platforms, and comparisons used.}
    \begin{tabular}{l l l l l p{2.7cm} c}
        \toprule
        & \textbf{Reference} & \textbf{Method/Novelty} & \textbf{Category} & \textbf{Simulation}  & \textbf{Comparison} & \textbf{Real World} \\
        \midrule
        \multirow{12}{*}{\rotatebox[origin=c]{90}{Classical}} 
            & O'Kelly et at. (2020) \cite{o2020tunercar, zheng2022gradient} & Genetic optimisation  & Offline planing &  F1TENTH Gym  & Gradient-based trajectory optimisation \cite{Heilmeier2020MinimumCar} & \cmark \\ 
            & Li et al. (2022) \cite{li2022real} & Multi-vehicle MPC & Online planning & Custom   & \xmark  & \xmark \\
            & Curinga (2018) \cite{curinga2017autonomous} & MPCC & Online planning & Custom  & Optimisation \& tracking  & \xmark \\
            & Bhargav et al. (2021) \cite{bhargav2021track} & MPCC for overtaking & Online planning & \Fone Gym   & \xmark  & \xmark \\
            
            & Jain et al. (2020) \cite{Jain2020BayesRace} & MPC with gaussian process & Control &  F1TENTH Gym  & MPC & \xmark\\
            & Cataffo et al. (2022) \cite{cataffo2022nonlinear} & MPC for obstacle avoidance & Control & F1Tenth.dev    & \xmark  & \xmark \\
            & Wang et al. (2021) \cite{wang2021deep} & Data-driven MPC & Control & Gazebo & MPC, adaptive pure pursuit & \cmark \\ 
            & Nagy et al. (2023) \cite{nagy2023ensemble} & Adaptive MPC & Control & F1TENTH Gym & Pure pursuit & \cmark	\\ 
            & Sukhil and Behl (2021) \cite{SukhilAdaptiveRacing} & Adaptive pure pursuit & Control & F1Tenth.dev   & Pure pursuit & \cmark \\
            & Becker et al. (2023) \cite{becker2023model} & Model- \& acceleration-based pure pursuit & Control & \xmark & Pure pursuit & \cmark	\\ 
            
            & Sezer et at. (2012) \cite{sezer2012novel} & Follow-the-gap & Mapless & Custom  & APF, A\textsuperscript{*} & \cmark \\
            & Otterness (2019) \cite{otterness2019disparity} & Disparity extender & Mapless & \xmark   & \xmark & \cmark \\

        \midrule

        \multirow{17}{*}{\rotatebox[origin=c]{90}{Learning}} 
            & Hamilton et al. (2022) \cite{hamilton2022zero}  & DRL vs IL & End-to-end  & Gazebo & IL, SAC, DDPG & \cmark \\ 
            & Brunnbauer et al. (2022) \cite{Brunnbauer2022LatentRacing} & Model-based DRL & End-to-end & PyBullet  & Model-free DRL &\cmark \\ 
            & Bosello et al. (2022) \cite{Bosello2022TrainRaces} & Generalisable DRL & End-to-end &  Rviz & DRL, Dreamer, FTG & \cmark \\
            & Sun et al. (2023)\cite{sun2023benchmark} & IL algorithm comparison & End-to-end &  F1TENTH Gym & IL, PPO, DAgger, EIL  & \cmark \\ 
            & Sun et al. (2023) \cite{sun2023mega} & New IL algorithm & End-to-end & F1TENTH Gym & IL, experts & \xmark \\
            & Evans et al. (2023) \cite{evans2023highspeed} & Trajectory-aided learning reward & End-to-end  & F1TENTH Gym  & Cross-track \& heading &  \xmark \\ 
            & Evans et al. (2023) \cite{evans2023comparing} & Comparing DRL architectures & Comparison & F1TENTH Gym & Optimisation \& tracking & \cmark \\ 
            
            & T\u{a}tulea-Codrean et al. (2020) \cite{tuatulea2020design}  & IL to replace MPC  & Learned planning &  Gazebo  &  MPC  & \xmark \\
            & Ghignone et al. (2022) \cite{ghignone2023tc} &  Trajectory-conditioned learning & Learned planning &  F1TENTH Gym  & MPC & \cmark \\
            & Dwivedi et al. (2022) \cite{dwivedi2022continuous} &  Plan-assisted DRL & Learned planning &  PyBullet  & Dreamer, MPO & \xmark \\
            & Trumpp et al. (2023) \cite{trumpp2023residual}  & RPL for high-speed racing & RPL &  F1TENTH Gym &  Optimisation \& tracking & \xmark	\\ 
            
            & Evans et al. (2021) \cite{evans2021learning, evans2021reward} & Static obstacle avoidance & RPL &  Custom & FTG, end-to-end DRL & \xmark\\
            & Zhang et al. (2022) \cite{zhang2022residual} & APF-based RPL & RPL & PyBullet & FTG, MPC, DRL, Dreamer & \xmark \\
            & Trumpp et al. (2024) \cite{trumpp2024racemop} & Mapless RPL racing  & RPL &  F1TENTH Gym  & APF, disparity extender & \xmark \\
            
            & Ivanov et al. (2020) \cite{ivanov2020case} & Formal neural network verification & Safe learning & Custom & DRL & \cmark \\
            & Musau et al. (2022) \cite{Musau2022OnControllers} & Reachability Theory safety & Safe learning & Gazebo & IL, DRL & \cmark \\
            & Evans et al. (2023) \cite{evans2023safe, evans2023bypassing}  & Viability Theory safety   & Safe learning &  F1TENTH Gym & DRL, pure pursuit & \cmark \\ 


         \bottomrule
    \end{tabular}
    \label{tab:literature_table}

\end{table*}

\textbf{Mapless Methods:}
A commonly used mapless approach for F1TENTH racing is the Follow-the-Gap (FTG) controller. Proposed by Sezer et at.~\cite{sezer2012novel}, this is a reactive method based on identifying gaps in the car's LiDAR signal. 
After identifying all gaps in the current LiDAR scan, the car's steering angle is directly defined as the angle in direction of the widest or furthest gap. An extension to FTG is the disparity extender \cite{otterness2019disparity}. 
In this approach, before detecting the gaps in the LiDAR signal, all disparities in the LiDAR signal are identified. The distance readings of the LiDAR signal adjacent in some distance to the disparity are set to the minimum distance in each interval. Typically, the interval is chosen as the vehicle width plus a safety margin, thus avoiding trajectories that cut obstacles closely.

\subsection{Learning Approaches} \label{subsec:lit_deep_learning}

Deep learning approaches use neural networks to replace a part or all of the racing pipeline.
We study learning approaches in the categories of architecture, algorithm and reward signal.
The deep learning architecture determines the role of the agent in the racing pipeline.
The algorithm denotes the process of training the neural network from random initialisation to a policy that controls the vehicle.
The reward signal is the method of communicating the desired behaviour to the agent.

\begin{figure}[t]
    \centering
    \includegraphics[width=\linewidth]{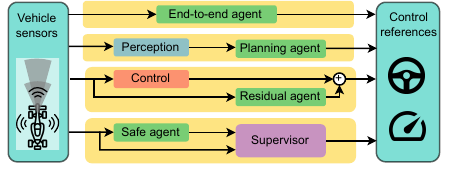}
    \caption{End-to-end, planning, residual, and safe agent architectures for autonomous racing.}
    \label{fig:racing_drl_setup}
\end{figure}

\textbf{Learning architectures:}
The learning architecture describes the input (state vector) given to the agent and the role/use of the actions generated.
Fig. \ref{fig:racing_drl_setup} shows the end-to-end, planning, residual and safe learning architectures for autonomous racing.
We consider approaches using each architecture.

\textit{End-to-end learning} uses a neural network to replace the entire racing pipeline.
In F1TENTH racing, common inputs for end-to-end agents are single or multiple downsampled 2D-LiDAR scans and the vehicle speed \cite{sun2023benchmark, Brunnbauer2022LatentRacing}. 
Many studies have addressed constant speed racing (driving) \cite{hamilton2022zero, evans2023bypassing}, and implemented their approaches on physical vehicles at low, constant speeds \cite{Bosello2022TrainRaces, Brunnbauer2022LatentRacing}.
End-to-end approaches generalise well to unseen tracks \cite{Bosello2022TrainRaces} and transfer to physical vehicles \cite{evans2023comparing}.
A major challenge for end-to-end learning agents is reliable performance at high speed  \cite{evans2023highspeed}.

\textit{Planning agents} learn a policy that incorporates the vehicle's location and upcoming track (usually in the form of centre points) into the state vector \cite{ghignone2023tc, dwivedi2022continuous}. 
Including the track points in the state has been shown to outperform end-to-end agents in training stability, sample efficient and racing performance \cite{evans2023comparing}.
Another motivation is to replace an online trajectory planner (MPC) with a neural network to reduce the computational burden  \cite{tuatulea2020design, LuizaNeuralVehicle}.
The limitation of planning agents is they require localisation, restricting them to mapped environments and putting them in direct competition with optimisation approaches.

\textit{Residual policy learning (RPL)} learns an additional policy that is added to a classical base policy \cite{trumpp2023residual}.
Residual policies for racing have been learned on top of pure pursuit controllers \cite{trumpp2023residual} and artificial potential fields \cite{zhang2022residual}, resulting in faster convergence and improved stability.
Residual policy learning has been extended to obstacle avoidance \cite{evans2021learning} and multi-agent overtaking manoeuvres \cite{trumpp2024racemop}.
Learning residuals promise to harness the flexibility of neural networks to improve upon classical components, but have yet to be deployed on physical vehicles.

\textit{Safe learning} is a method for training neural network controllers while simultaneously ensuring vehicle safety \cite{evans2023safe}.
Safe learning approaches use a supervisor (action mask) to ensure that the controls executed on the vehicle are safe.
Actions masks have used formal methods \cite{ivanov2020case}, reachability theory \cite{Musau2022OnControllers}, time-to-collision (TTC) \cite{Bosello2022TrainRaces} and Viability Theory \cite{evans2023safe} to select which actions are safe.
Safe learning enables safe deployment on physical hardware, resulting in many approaches having physical evaluations \cite{ivanov2020case, Bosello2022TrainRaces, Musau2022OnControllers, evans2023bypassing}.
A challenge in safe learning is building action masks that accurately discern safe and unsafe actions for high-performance systems.

\textbf{Training algorithms:}
While the majority of learning approaches are trained with reinforcement learning (RL) \cite{Bosello2022TrainRaces, evans2023comparing, ghignone2023tc}, imitation learning (IL) has also been used \cite{sun2023benchmark, hamilton2022zero, sun2023mega, zhang2023f1tenth}.
Reinforcement learning trains agents to control a vehicle in an environment from the experience of receiving a state, selecting an action and receiving a reward.
In contrast, imitation learning trains the network to mimic the behaviour of an expert, as recorded in a dataset.
A study comparing the two found that RL is more robust to unseen states \cite{hamilton2022zero}, a result backed up by the poor completion rates shown by IL algorithms in \cite{sun2023benchmark}.

For training agents, the most common DRL algorithms have been TD3 \cite{evans2023highspeed, ivanov2020case}, SAC \cite{ghignone2023tc, hamilton2022zero}, and PPO \cite{trumpp2023residual, zhang2022residual}.
Other algorithms that have been used are the deep-Q-networks (DQN) \cite{Bosello2022TrainRaces}, and Dreamer, a model-based algorithm \cite{Brunnbauer2022LatentRacing}.
TD3 has shown to outperform SAC and DDPG for continuous control in autonomous racing \cite{evans2023comparing}.
It remains to be studied how off-policy algorithms (TD3) compare to on-policy algorithms (PPO) for autonomous racing.

\textbf{Rewards signals:}
There has been much variation in the reward signals used to train DRL agents for autonomous racing.
The reward signal plays a crucial role in the training process, but is often simply listed as a minor implementation detail.
One of the most common is the progress reward signal, which rewards vehicles according to their progress along the track \cite{Brunnbauer2022LatentRacing}. 
However, due to its lack of ability to communicate speed information, rewards that include the vehicle speed, and punish lateral deviation from the centre line have been favoured \cite{evans2021reward, Bosello2022TrainRaces}.
A trajectory-aided learning (TAL) reward has been proposed that uses an optimal trajectory to train the agent \cite{evans2023highspeed}.
Reward signal design for autonomous racing is difficult because the agent must be encouraged to race at high-speeds, while not being so aggressive that the vehicle crashes.

\subsection{Literature Summary}

Table \ref{tab:literature_table} presents a study of approaches to F1TENTH racing by identifying their category, evaluation benchmarks and if tested on a physical vehicle.
Since gradient-based offline trajectory optimisation approaches work well in \Fone racing, few studies have designed such approaches specifically for the \Fone platform.
Approaches in online planning lack comparison with other methods and evaluation on physical vehicles. 
This is possibly due to the complexity of the method and the computational burden, making them poorly suited to real-time embedded platforms.
In control, the challenge of high-speed tracking, taking the vehicle's non-linear dynamics into account, has been attempted with data-driven MPC \cite{nagy2023ensemble, wang2021deep}, and the model- and acceleration-based pure pursuit algorithm \cite{becker2023model}.
However, it is difficult to compare these approaches due to the numerous design decisions (model choice, track, horizon length) that differ between papers, and work is required to compare these methods directly.

There are many approaches in all learning categories for F1TENTH racing, with end-to-end being the largest.
While learned planning has shown to be more robust \cite{evans2023comparing}, end-to-end methods have been tested on physical platforms a lot more \cite{hamilton2022zero, Brunnbauer2022LatentRacing}.
Few papers compare classical and learning-based methods, demonstrating the field's fragmentation.
It is proposed that this is due to differing motivations for different methods; classical methods aim for high performance \cite{Jain2020BayesRace}, while learning methods aim for generality \cite{Bosello2022TrainRaces}.
Several simulators have been used, such as PyBullet \cite{Brunnbauer2022LatentRacing}, Gazebo \cite{tuatulea2020design}, and F1Tenth.dev (which is a Gazebo-based simulator) \cite{Babu2020F1tenthDevSimulator}.
A positive trend is the widespread adoption of the \Fone Gym simulator, allowing for results to be compared as demonstrated by \cite{Bosello2022TrainRaces}.

Classical racing approaches using estimation, optimisation and control have proven highly effective in many contexts, offering both high performance and safety guarantees.
However, their reliance on accurate dynamics models and apriori track knowledge restricts their applicability in unmodelled and unmapped contexts.
In contrast, the follow-the-gap method requires neither a map nor a dynamics model but often compromises performance.
End-to-end reinforcement learning agents excel in unmapped settings, allowing neural networks to learn generalisable racing policies.
Yet, concerns about the robustness and the complexity of their training pipelines persist.
Recent efforts to combine the adaptability of neural networks with the reliability of classical methods, such as those seen in planning, residual, and safe learning agents, show promise but have yet to achieve widespread adoption.

\section{Racing Methods} \label{sec:racing_methods}

Currently, many methods for driving and racing autonomously with the F1TENTH vehicle are available. 
To unify this, we describe here the most common racing algorithms from each category to provide an overview of different solution approaches.
We illustrate the particle filter algorithm used for localisation, a trajectory planning and control strategy, the model predictive contouring algorithm, the follow-the-gap method and end-to-end deep reinforcement learning.

\subsection{Particle Filter Algorithm}

The particle filter (PF) algorithm uses the control input and the 2D-LiDAR scan to estimate the vehicle's pose on the map \cite{WalshCDDT:Localization}.
The algorithm is a recursive Bayesian state estimator designed for non-parametric distributions by representing the belief of the vehicle's pose by a set of particles \cite{thrun2005probabilistic}.

\begin{figure}[t]
    \centering
    \includegraphics[width=\linewidth]{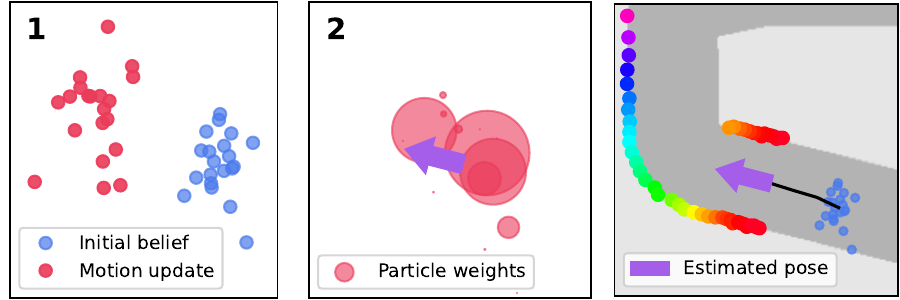}
    \caption{Motion and measurement updates for the particle filter.}
    \label{fig:particle_filter_plot}
\end{figure}

Fig. \ref{fig:particle_filter_plot} shows the particle filter process for a vehicle in the environment shown on the right.
A set of particles (blue) represents the belief of the vehicle's location.
The motion update uses the control input and a vehicle model are used to update the particles to form the posterior belief (red).
After each measurement from the LiDAR scan, weights (orange) are calculated for each particle based on the probability of the measurement being correct.
The current pose (purple arrow) is estimated by multiplying the particle locations with the weights.

\subsection{Trajectory Optimisation and Tracking} \label{subsec:two_stage}

We consider a two-stage planner that uses the trajectory optimisation method presented by \cite{Heilmeier2020MinimumCar} and the pure pursuit path tracking algorithm from \cite{coulter1992implementation}.
The two-stage nature means a trajectory can be calculated offline and tracked during the race.

\begin{figure}[t]
    \centering
    \includegraphics[width=0.7\linewidth]{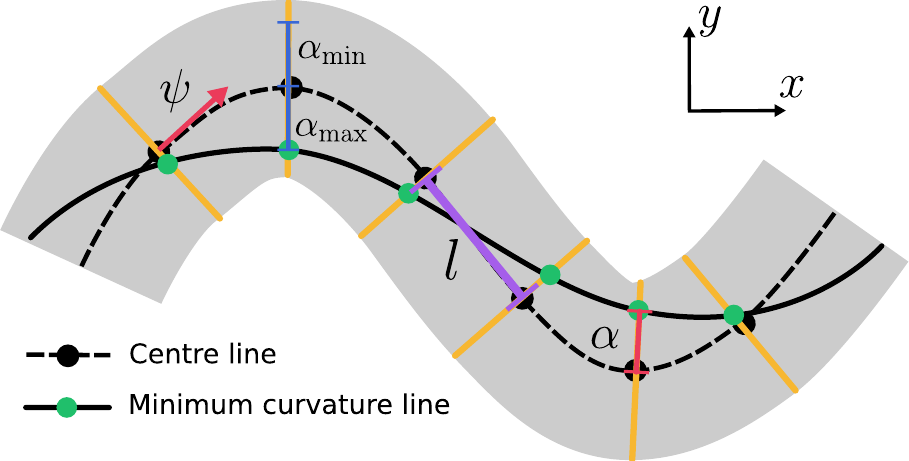}
    \caption{Centre line, minimum curvature line, heading angle $\psi$, segment length $L$, and optimisation variable $\alpha$ and limits $[\alpha_{\text{min}}, \alpha_{\text{max}}]$ for a track segment.}
    \label{fig:OptimisationTrackMap}
\end{figure}

\textbf{Trajectory optimisation:}
The input into the optimisation is a list of centre line points and track widths from the map.
Fig. \ref{fig:OptimisationTrackMap} shows a track segment indicating the centre-line points used to calculate the heading angle $\psi$ and segment length $l$.
The minimum curvature path generation minimises the change in heading angle over the segment length.

The optimisation variable is $\alpha$, which is the (positive or negative) distance from the track centre measured along the normal vector (perpendicular to the heading angle).
The minimum and maximum values for $\alpha$ are defined by the track widths.
The optimisation can be written as,
\begin{align}
    \text{minimise}_{[\alpha_1...\alpha_N]} \quad &  \quad\sum_{i=1}^N \bigg( \frac{d \psi}{dl} \bigg) ^2 \\
    \text{subject to} \quad & \quad \alpha_i \in [\alpha_{i, \text{min}}, \alpha_{i, \text{max}}] 
\end{align}

The minimum time speed profile is calculated as the maximum speed that keeps the vehicle within the handling limits.
For each point on the path $i$, the largest speed for the next point $v_\text{i+1}$ is calculated as, 
\begin{equation}
    v_{i+1} = \sqrt{v_i^2 \pm 2 l_i \cdot \mu a_\text{max}(1-v_i \kappa_i)}.
\end{equation}
The inputs into the calculation are the maximum acceleration (based on the motors), current speed $v_i$, segment length $l_i$, maximum vehicle acceleration $a_\text{max}$, path curvature $\kappa_i$ and friction coefficient $\mu$.
The $\pm$ indicates that the process is repeated in both forward and reverse path directions.

\textbf{Pure pursuit:}
The pure pursuit path tracking algorithm tracks a reference path using a lookahead distance to select a point to steer towards.
Given a lookahead point that is a lookahead distance $l_\text{d}$ away from the vehicle, and at a relative heading angle of $\phi$, the steering angle $\delta$ is calculated as, 
\begin{equation}
    \delta = \arctan \bigg( \frac{L \sin (\phi)}{l_\text{d}} \bigg),
\end{equation}
where $L$ is the length of the vehicle's wheelbase.
The lookahead distance $l_\text{d}$ has constant and speed-dependant parts.
Additionally, we limit the speed based on the steering angle to prevent the algorithm from selecting a large steering angle if it is going too quickly.
The speed action is the minimum of the raceline speed $v_\text{raceline}$ and the speed that reaches a lateral force to weight ratio of 1.5, calculated as,
\begin{equation}
    v = \min (v_\text{raceline}, \sqrt{1.5 \cdot g \cdot L / \tan(|\delta|)}).
\end{equation}


\subsection{Model Predictive Contouring Control}

Model predictive control is a receding horizon strategy that calculates a set of control inputs that optimise an objective function subject to constraints \cite{Liniger2015OptimizationCars}.
Model predictive contouring control (MPCC) maximises the progress along the centre line subject to the path being on the track and dynamically feasible.

The optimisation variables consist of the state and control inputs at each discrete timestep.
The state is represented by position $x, y$ and heading $\theta$, and the control actions are speed $v$ and steering $\delta$.
The state of centre line progress $s$ and corresponding control of centre line speed $\dot s$ are added, resulting in a 4D state $[x, y, \theta, s]$ and 3D control $[\delta, v, \dot s]$.
The initial state is set to the vehicle's position, and subsequent states are constrained to be the result of the control actions according to the kinematic bicycle model.
The friction is constrained by ensuring that the lateral force is smaller than the friction coefficient $\mu$ multiplied by the vehicle's weight (mass $m$ times gravitational force $g$) as,
\begin{equation}
    v^2 /L \cdot \tan(|\delta|)m < \mu \cdot mg.
\end{equation}

\begin{figure}[t]
    \centering
    \includegraphics[width=0.9\linewidth]{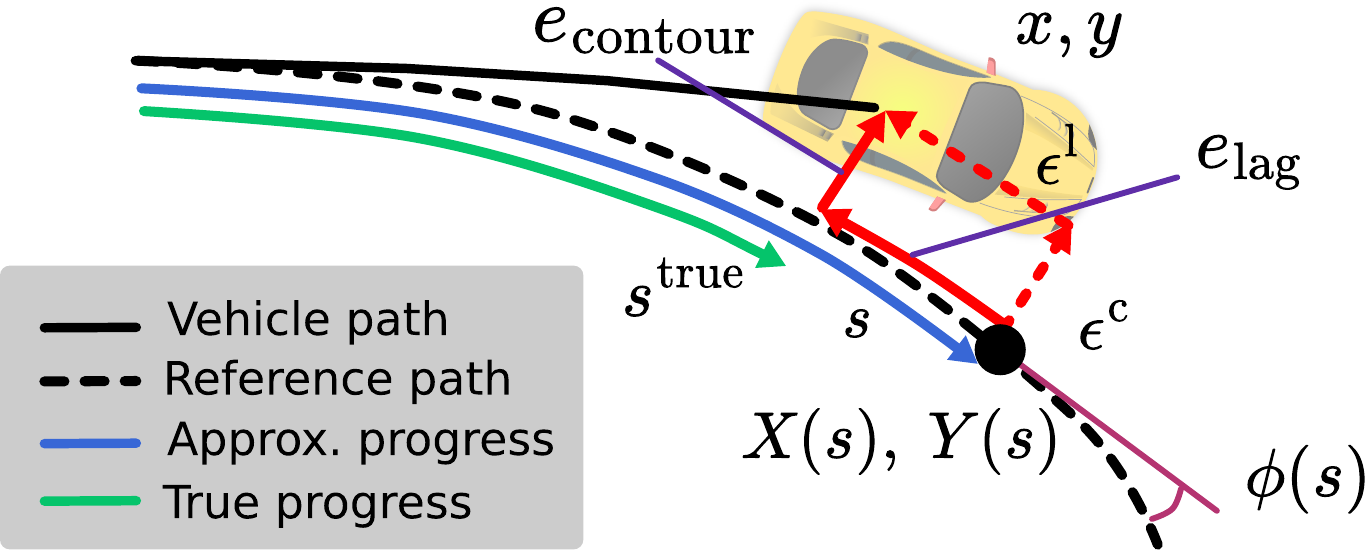}
    \caption{Lag and contouring errors represent vehicle progress along the path.}
    \label{fig:mpcc_errors}
\end{figure}

\textbf{Objective:}
The MPCC objective maximises progress along the path using four terms: contouring error, lag error, progress, and control regularisation.
Fig. \ref{fig:mpcc_errors} shows how the contouring and lag errors are measured.

The $x$ and $y$ coordinates and heading angle of the track are represented as functions of progress along the centre line path $X(s),~ Y(s),~ \Phi(s)$.
Since the true progress along the reference path $s^\text{true}$ cannot be found, an approximate progress $s$ is used.
Fig. \ref{fig:mpcc_errors} shows how the lag error quantifies the difference between the true and approximate progress.
The contouring error quantifies the vehicle's perpendicular distance from the trajectory.
For the state $(x, y, \theta, s)$, the linearised contouring and lag errors ($\epsilon^c$ and $\epsilon^l$) can be calculated as,
\begin{align}
    \epsilon^\text{c} & = \sin(\Phi(s)) (x - X(s)) - \cos (\Phi(s))(y-Y(s)) \\
    \epsilon^\text{l} & = -\cos(\Phi(s)) (x - X(s)) - \sin (\Phi(s))(y-Y(s))
\end{align}
The progress speed along the centre line is rewarded to encourage the vehicle to maximise progress along the track.
Finally, a control regularisation term penalises steering angles, encouraging smooth behaviour.
Each term is weighted, and the weights are tuned to ensure satisfactory performance.


\subsection{Follow-the-gap Method}

The follow-the-gap (FTG) method, extended to the disparity extender algorithm, is a simple, mapless approach to autonomous racing \cite{sezer2012novel, otterness2019disparity}.
The algorithm identifies the nearest boundary/obstacle in the LiDAR scan and excludes a bubble of the beams around it.
Then, the largest visible gap in the LiDAR scan is identified, and the steering angle is calculated to drive towards the middle of the gap.
The speed is selected using a higher speed (5 m/s) for small steering angles and a low speed (3 m/s) for high steering angles.

\subsection{End-to-end Deep Reinforcement Learning}  \label{subsec:learning_racing_methods}

End-to-end learning agents for autonomous racing use a neural network to map a state vector $s$ directly to a control action $a$ \cite{evans2023comparing}.
Mathematically, the mapping is written as a policy $\pi$ that maps a state to an action as $a = \pi(s)$.
The policy is a deep neural network of multiple fully connected layers.
The state vector comprises the previous and current LiDAR scans and the vehicle speed.
The control actions are the speed and steering angle used to control the vehicle.

\begin{figure}[t]
    \centering
    \includegraphics[width=\linewidth]{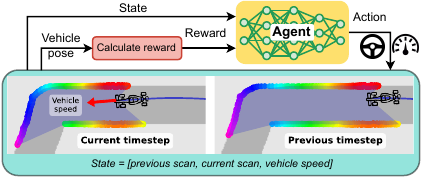}
    \caption{Traininng setup for training a reinforcement learning agent to race in a simulator.}
    \label{fig:drl_training_setup}
\end{figure}

Fig. \ref{fig:drl_training_setup} shows the training setup for an end-to-end reinforcement learning agent.
During training, the agent experiences states, selects actions and receives rewards.
The reward, which indicates how good or bad an action is, is calculated based on the vehicle's pose and/or the agent's action.
The reinforcement learning objective is to select actions that maximise the total reward \cite{sutton2018reinforcement}.
The training is split into episodes where the vehicle is randomly spawned on the track and must drive until it either crashes or completes a lap.

We use the TD3 algorithm \cite{fujimoto2018addressing}, a continuous control, off-policy, actor-critic reinforcement learning algorithm.
The actor network is the policy that selects actions, and the critic network learns a Q-value function for each state-action pair.
The critic is trained towards targets calculated by the Bellman update equation. 
The actor is updated using the policy gradient that maximises Q-values.

\begin{figure}[t]
    \centering
    \includegraphics[width=\linewidth]{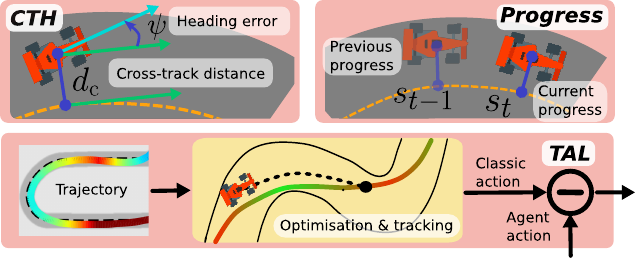}
    \caption{Illustrations of the cross-track and heading, track progress and trajectory-aided learning reward signals.}
    \label{fig:reward_signals}
\end{figure}

To train the agent to complete laps without crashing, the agent is rewarded with 1 for completing a lap, punished with -1 if the vehicle crashes, and given a shaped reward to encourage fast racing behaviour.
Fig. \ref{fig:reward_signals} illustrates the three shaped reward signals of cross-track and heading error (CTH), track progress and trajectory-aided learning (TAL).

The \textit{cross-track and heading error (CTH)} reward punishes the cross-track distance $d_\text{c}$ and promotes speed in the direction of the track.
The normalised vehicle speed $v_t$, heading error $\psi$ and cross-track error $d_c$ are used to calculate the reward as,
\begin{equation}
    r_\text{CTH}  = v_\text{t}\cos \psi -  ~ d_\text{c},
\end{equation}

The \textit{track progress} reward promotes advancement along the track centre line by rewarding the progress made during each timestep.
The current progress $s_t$ and previous progress $s_{t-1}$ are used to calculate the reward as,
\begin{equation}
      r_\text{progress} = s_{t} - s_{t-1},
\end{equation}

The \textit{trajectory-aided learning (TAL)} reward trains the agent to mimic the actions selected by the optimisation and tracking planner described in \S \ref{subsec:two_stage} \cite{evans2023highspeed}.
The agent's action $\mathbf{u}_\text{agent}$ and classical action (optimisation and tracking) $\mathbf{u}_\text{classic}$ is used to calculate the TAL reward as,
\begin{equation}
        r_\text{TAL} =  1 - | \mathbf{u}_\text{agent} - \mathbf{u}_\text{classic}  |
\end{equation}

\section{Benchmark Evaluation} \label{sec:evaluation}

We benchmark the results described in Section \ref{sec:racing_methods} by,
\begin{enumerate}
    \item Investigating the impact of localisation error and control frequency on racing performance
    \item Comparing the CTH, progress and TAL reward signals, and training maps on agent performance
    \item Providing benchmark results for future studies
\end{enumerate}
For each test, we present the key quantitative result (i.e. lap time) and then qualitative data (i.e. speed profile plots) to explain the reasons for the result.

\begin{figure}[t]
    \centering
    \setlength{\tabcolsep}{5pt}
    \begin{tabular}{cccc}
      \raisebox{0.2\height}{\includegraphics[width=0.95cm]{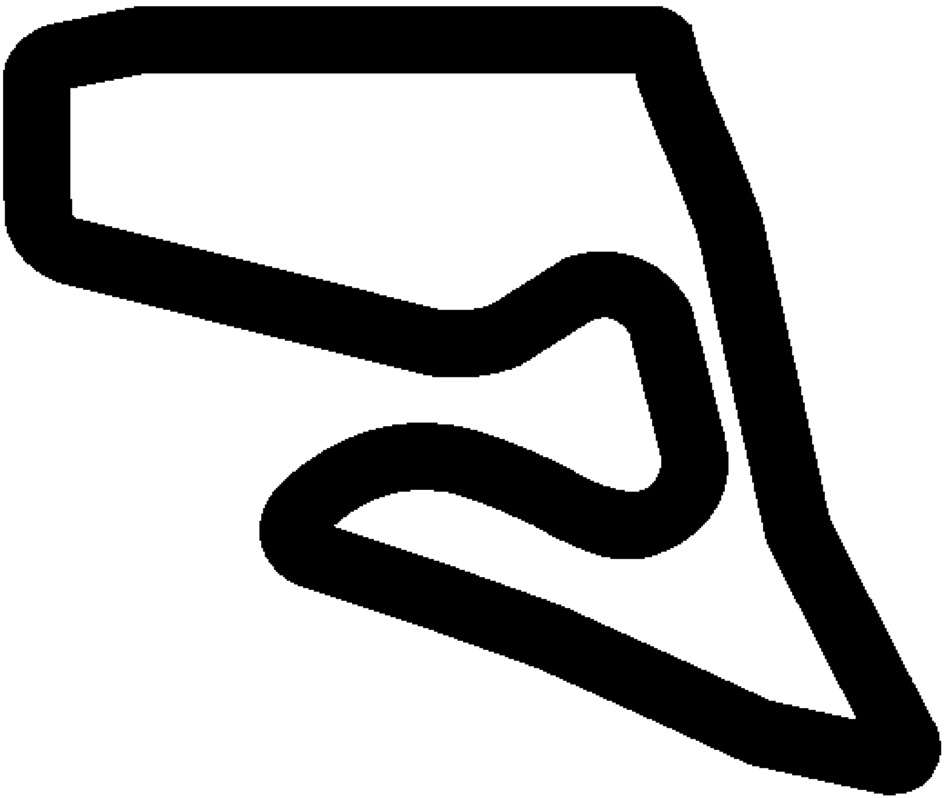}}  & \raisebox{0.2\height}{\includegraphics[width=2.4cm]{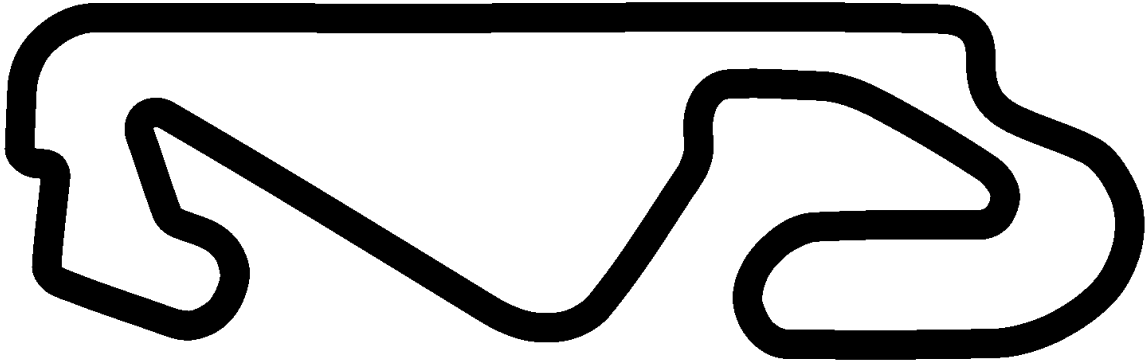}}   & \includegraphics[width=2.1cm]{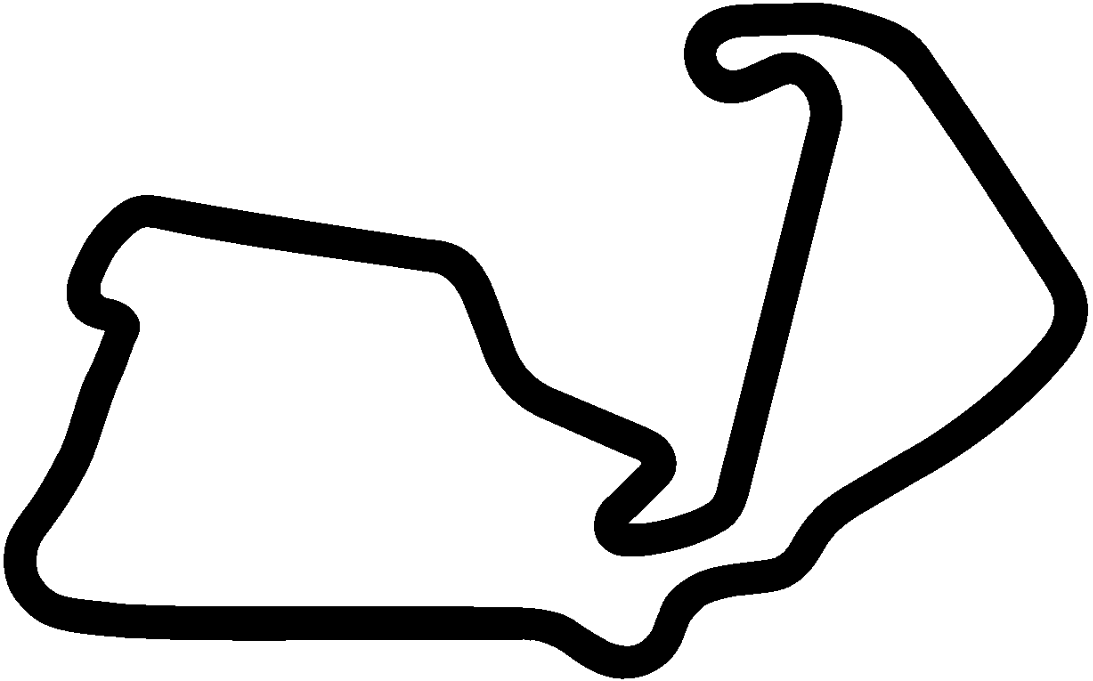} & \includegraphics[width=1.7cm]{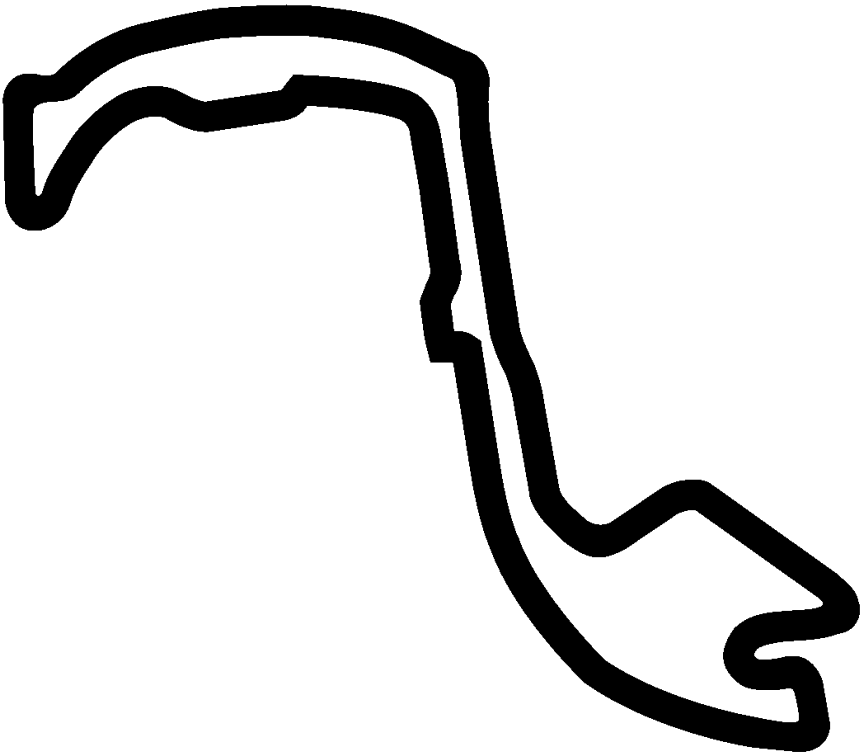} \\
    \end{tabular}
    \caption{Maps of the AUT, ESP, GBR, and MCO race tracks (left to right).}
    \label{fig:map_description}
\end{figure}

\textbf{Methodology:}
We use the open-source F1TENTH autonomous vehicle simulator \cite{Okelly2020F1TENTH_evaluation}.
The vehicle is represented by the single-track bicycle model (described in Appendix \ref{app:dynamic_model}), which describes the state in terms of position, speed, steering angle, yaw, yaw rate, and slip angle \cite{althoff2017commonroad}.
The 2D-LiDAR scanner is simulated with a ray-casting algorithm that measures the distance to the track boundary for each beam.
The simulator and all the tests are written in Python and run on an Ubuntu computer.
The simulator dynamics are updated at a rate of 100 Hz in all tests, and unless otherwise stated, the planning frequency is 25 Hz.

We use a set of four F1 race tracks that have been rescaled ($\approx$ 1:20) for F1TENTH racing and had the widths adjusted to a constant value.
These maps have been widely used in the literature \cite{evans2023comparing, Bosello2022TrainRaces, Brunnbauer2022LatentRacing}, since they provide a difficult challenge for F1TENTH vehicles and allow for results to be easily compared.
Fig. \ref{fig:map_description} shows the shape of the circuits in Austria (AUT), Spain (ESP), Great Britain (GBR) and Monaco (MCO).
Appendix \ref{app:map_information} provides the track length, percentage of straight sections and number of corners for each track.

For all tests, 10 laps are run with starting points randomly selected along the centre line.
All the results are seeded for reproducibility.
The following metrics are used,
\begin{itemize}
    \item Lap time: the mean of the lap times for successfully completed laps.
    \item Completion rate: the number of successfully completed laps divided by the total number of test laps
    \item Progress: the progress made by the vehicle before crashing (used for learning tests)
\end{itemize}
The speed and slip angle profiles and the path taken are analysed to better understand the results by exposing the vehicle behaviour.

\subsection{Localisation Error Evaluation}


Using a particle filter for pose estimation introduces error into the racing pipeline that many planning and control solutions ignore.
Therefore, we investigate the impact of localisation by recording the lap times using the true pose (from the simulator) and the estimated pose.
The estimated pose comes from the particle filter, which was tuned to have 1000 particles (see Appendix \ref{app:particle_filter_tuning}).
We test laps on the AUT track with friction values ranging from 0.5 to 1.0.

\begin{figure}[t]
    \centering
    \includegraphics[width=\linewidth]{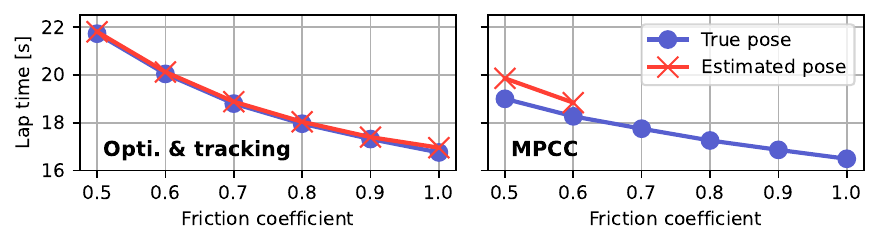}
    \vspace{-3mm}
    \caption{Lap times using estimated and true friction values on AUT map.}
    \label{fig:FrictionTimeComparison}
\end{figure}

\begin{figure}[t]
    \centering
    \includegraphics[width=\linewidth]{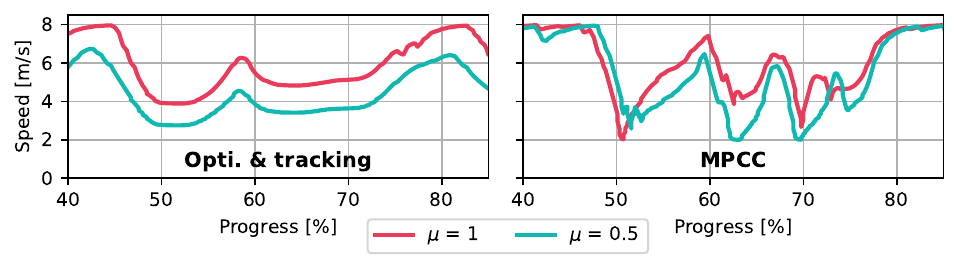}
    \caption{Speed profiles on AUT using friction coefficients of 0.5 and 1.0}
    \label{fig:speed_friction_comparison_aut}
\end{figure}

Fig. \ref{fig:FrictionTimeComparison} shows the lap times for the optimisation and tracking and MPCC planners using the estimated and true poses with varying friction values.
The optimisation and tracking planner completes laps at all the friction values.
The MPCC planner using the true pose can complete laps at all friction values, but using the estimated pose, only up to 0.6.
The range of the optimisation and tracking planner lap times varies from 22 s to 17 s, while the MPCC planner from 19 s to 17 s.
Fig. \ref{fig:speed_friction_comparison_aut} qualitatively explains this difference using segments of the speed profiles with friction coefficients of 0.5 and 1.0.
The difference between the optimisation and tracking planner has a large, near constant speed difference.
Conversely, the MPCC has a smaller, irregular gap.
This result demonstrates that the MPCC planner is more affected by errors in the pose than the optimisation and tracking planner because it selects higher speeds, even at lower friction values.

\begin{figure}[t]
    \centering
    \includegraphics[width=\linewidth]{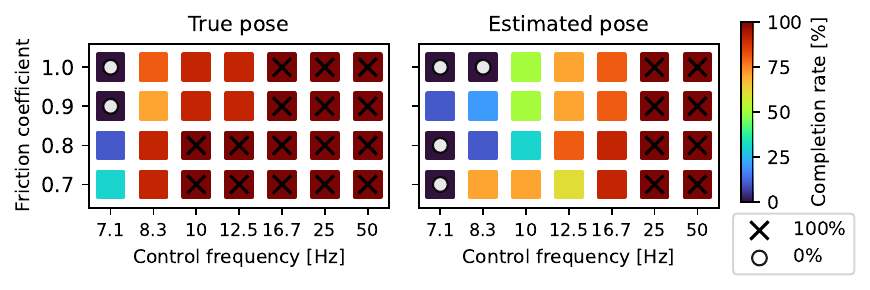}
    \caption{Friction and control frequency plotted against completion rate for the optimisation and tracking planner using the estimated and the true poses.}
    \label{fig:FrequencyFrictionCompletionRate}
\end{figure}

We investigate how control frequency impacts the effect of localisation using the particle filter with 200 particles (see Appendix \ref{app:particle_filter_tuning}) to expose the impact of the error.
Fig. \ref{fig:FrequencyFrictionCompletionRate} shows the optimisation and tracking planner completion rate, tested on the AUT map with varying friction coefficients and control frequencies.
Using the true pose results in the planner achieving a 100\% completion rate for many tests, even at lower control frequencies.
In contrast, using the estimated pose results in much lower completion rates, especially at high friction coefficients.

\textbf{Conclusion 1:}
Localisation error plays a significant role in lap times and completion rates, and therefore, tests should explicitly study how perception error and control frequency affect a planner's performance.

\subsection{Training Configuration Evaluation}

The learning evaluation compares the progress, CTH and TAL reward signals and investigates the impact of the training map on DRL agent performance.
Three agents are trained for each reward signal on each map for 50,000 steps.

\begin{figure}[t]
    \centering
    \includegraphics[width=\linewidth]{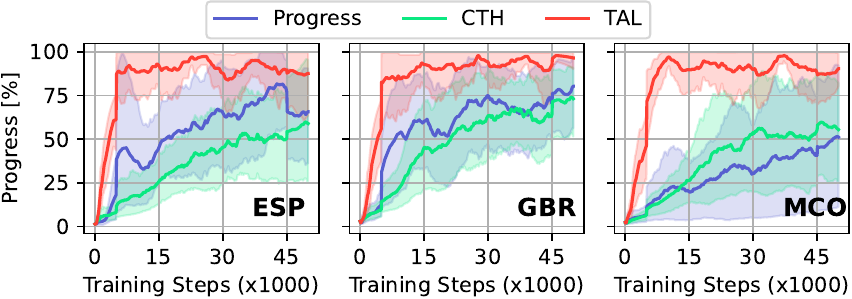}
    \caption{Average progress during training on the AUT, GBR and MCO tracks.}
    \label{fig:BenchmarkProgressTraining}
\end{figure}

Fig. \ref{fig:BenchmarkProgressTraining} shows the mean lap progress during training, with the shaded regions indicating the minimum and maximum.
The TAL reward produces the fastest training, and the agents reach around 90\% average progress in around 15k steps.
The CTH reward produces slower training, and the agents converge to around 70\% average progress.
The progress reward results in the agents converging to low average progresses, around 50\%.

\begin{figure}[t]
    \centering
    \includegraphics[width=\linewidth]{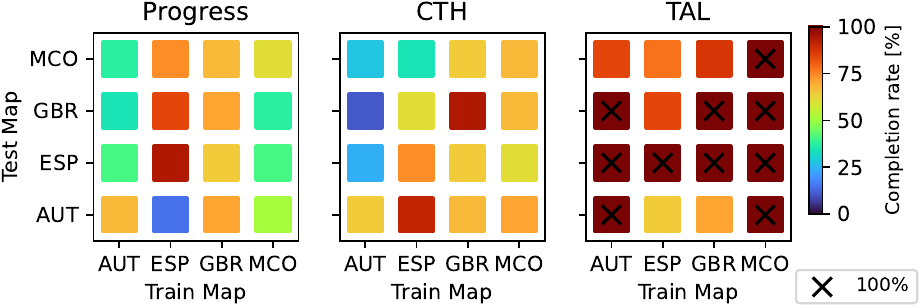}
    \caption{Completion rate for DRL agents tested on each map.}
    \label{fig:drl_CompletionRate}
\end{figure}

Fig. \ref{fig:drl_CompletionRate} shows the completion rate for each agent on each test map.
The progress agent achieves low completion rates on all the maps.
The CTH agents achieve higher completion rates, with some agents achieving around 80\%.
The TAL agents achieve the highest completion rates, with 10 of the tests achieving a 100\% completion rate.
The MCO map produces the best generalisation, with all the agents achieving 100\% on all test maps.

\begin{figure}[t]
    \centering
    \includegraphics[width=0.99\linewidth]{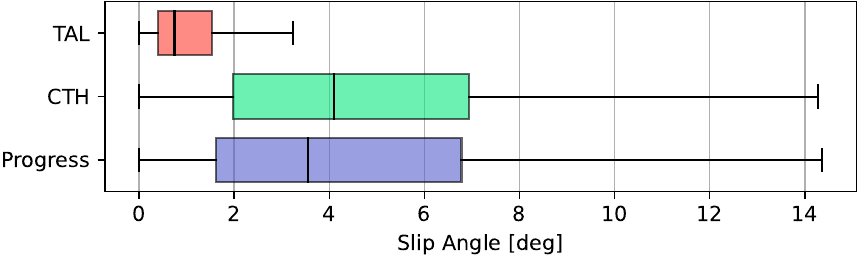}
    \caption{Box and whisker plot of the slip angles for a lap on the MCO track.}
    \label{fig:drl_slip_box_whisker_mco}
\end{figure}

Fig. \ref{fig:drl_slip_box_whisker_mco} shows a box and whisker plot of the slip angles for a lap on the MCO track. 
The progress and CTH rewards both have median slip angles of around 4 degrees and maximums of over 14 degrees.
These are much higher than the TAL's median of 18 degrees and a maximum of 5 degrees.
This suggests that the TAL reward outperforms the CTH and progress reward signals due to its trajectories having lower slip angles \cite{evans2023highspeed}.

\textbf{Conclusion 2:}
The TAL reward has the highest completion rate and produces the lowest slip angles.

\subsection{Benchmark Results}

We present benchmark results by providing lap times and supporting speed profiles.
We record the average lap times of the optimisation and tracking, MPCC, follow-the-gap and end-to-end planners.
The classical planners use a friction coefficient of 0.9 and the true pose from the simulator. 
The end-to-end agent is trained on GBR with the TAL reward.

\begin{table}[t]
    \renewcommand{\arraystretch}{1.4}
    \begin{tabular}{w{l}{2.1cm}  w{c}{1.1cm} w{c}{1.1cm} w{c}{1.1cm} w{c}{1.1cm}}
        \toprule
        \multirow{2}{*}{\textbf{Planner}} & \multicolumn{4}{c}{\textbf{Map}} \\
         & AUT & ESP & GBR & MCO \\
        \midrule
        Opti. \& tracking & \textbf{16.79} & \textbf{35.92} & \textbf{31.24} & \textbf{28.08} \\
        MPCC & 16.87 & 39.13 & 35.40 & 31.53 \\
        Follow-the-gap & 19.10 & 45.78 & 39.34 & 34.99 \\
        End-to-end & 19.94 & 46.37 & 40.22 & 34.93 \\
        \bottomrule
    \end{tabular}
    \vspace{2mm}
    \caption{Mean lap times [s] for the optimisation and tracking, MPCC, follow-the-gap and end-to-end methods.}
    \label{tab:bench_lap_times}
\end{table}

Table \ref{tab:bench_lap_times} presents the mean lap times for the planners on the four test maps.
The optimisation and tracking planner achieved the fastest lap time on all the maps.
The MPCC is 0.1 s slower on the AUT map and ranges to 4.16 s slower on GBR.
The follow-the-gap and end-to-end methods achieve similar lap times, significantly slower than the other two approaches.

\begin{figure}[t]
    \centering
    \includegraphics[width=\linewidth]{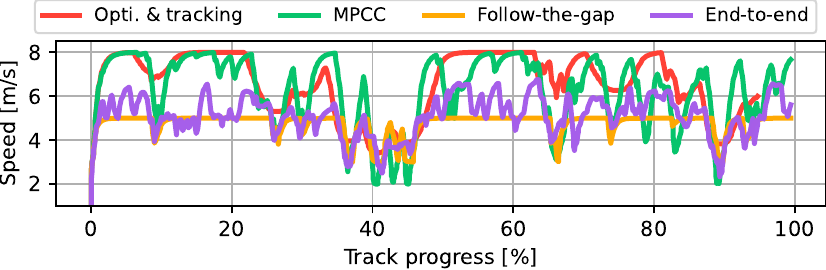}
    \caption{Comparison of the speed profiles for the AUT map.}
    \label{fig:speed_comparison_aut}
\end{figure}

Fig. \ref{fig:speed_comparison_aut} shows the speed profiles for each planner on the AUT map.
Both classical planners select a range of speeds, slowing down at times (around 40\%) and seeing up to the maximum speed.
The follow-the-gap planner has a maximum speed of 5 m/s and the end-to-end agent only selects speeds of above 6 m/s for short periods before slowing down again.

\begin{figure}[!t]
    \centering
    \includegraphics[width=\linewidth]{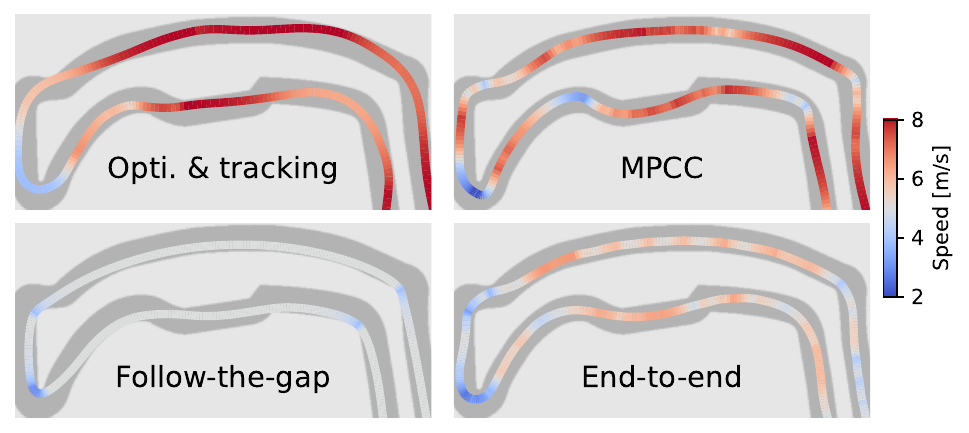}
    \caption{Trajectories with colour indicating speed on the MCO track}
    \label{fig:speed_trajectories}
\end{figure}

Fig. \ref{fig:speed_trajectories} further explains the behaviour by showing trajectory segments for each planner on the MCO map.
The optimisation and tracking planner selects a smooth profile that slows down in the corners and speeds up in the straighter sections.
The MPCC planner selects a more jagged speed profile by slowing down more in the corners.
The follow-the-gap method often drives in a straight line and always cuts to the inside of the corner while slowing down.
The end-to-end planner sometimes speeds up, but not to the full speed, and selects a varying speed profile that appears to change for no reason.



\textbf{Conclusion 3:}
Offline trajectory optimisation and tracking achieves the fastest lap times for F1TENTH autonomous racing.
It is followed by MPCC, with the follow-the-gap method and end-to-end agents being significantly slower.

\section{Conclusion} \label{sec:conclusion}


This paper addressed the fragmented field of F1TENTH autonomous racing.
The literature study described how different methods have been used for the subproblems of classical and learning-based approaches.
The study explained how the methods relate to one another, creating a taxonomy for future work to fit into. 
The methods of particle filter localisation, trajectory optimisation and tracking, model predictive contouring control, follow-the-gap and end-to-end deep reinforcement learning were described.
The evaluation highlighted the importance of reporting on the overlooked factors of localisation error for classical approaches and reward signals for learning methods.
The benchmark evaluation concluded that trajectory optimisation and tracking solutions provide the fastest lap times due to optimal speed profile selection.

\subsection{Future Work}

\textbf{Vision-based racing:}
Almost all racing methods use LiDAR as the main input modality.
While LiDAR's are convenient since they provide geometric information about the surrounding environment, they are expensive and poorly suited to real-world applications.
In contrast, cameras are cheaper and contain more expressive information.
Therefore, using cameras for high-performance control should be investigated.
Cai et al. \cite{cai2021vision} started work in this direction with end-to-end neural networks; however, their solution is limited to slow speeds and neglects the key racing challenges.

\textbf{3D-Aware Racing:}
While the works in this paper focus on racing in 2-dimensions the world exists in 3-dimensions. 
An interesting initial direction is 3D simulators that can accurately simulate 3D-LiDAR and visual camera input.
Additional research areas are 3D mapping, object detection and localisation.
Solutions focusing on 3-dimensions will improve the applicability of many methods to real-world robotics and autonomous driving pipelines.

\textbf{Full-stack racing:}
Many approaches have focused on a specific section of the planning pipeline, made possible due to solution modularity.
However, further research is required to investigate the effect of one section on another.
For example, approaches focusing on planning, should consider the impact of noisy, delayed localisation information.
A key point is that evaluations should include full-stack performance in addition to submodule comparison tests.

\textbf{Robustness of end-to-end learning:}
End-to-end learning has shown excellent potential in various games such as Grand Turismo Sport \cite{Wurman2022OutracingLearning}.
However, these have not yet been seen in real-world robotics tasks due to a lack of robustness one end-to-end solutions, i.e. many agents still occasionally crash.
This limitation prevents the widespread use of end-to-end neural networks.
Work in autonomous racing and other robotics fields should address these challenges and design more robust algorithms with safety guarantees.

\textbf{Simulation-to-reality transfer:}
The literature study showed that many approaches are only tested in simulation. 
While simulation is helpful since you can easily compare methods, it neglects the true noisy dynamics of physical systems.
Therefore, studies should focus on the differences between simulation and real-world results.
This is specifically needed in the machine learning approaches, where simulation testing is common.

\textbf{Mapless solutions:}
A majority of approaches, especially in the categories of optimisation-based planning and control, assume that a map of the environment is available.
While this is often the case in racing, it is not the case in many real-world applications.
Therefore, research should investigate solutions that do not require a map, and thus promise flexibility.
This will promote flexible, adaptable, general solutions that can be used in other robotics applications.

\textbf{Multi-agent racing:}
The crown of racing is multi-vehicle, competitive racing, where software stacks compete against each other.
Competitive, multi-vehicle racing, similar to real-world events like F1, is a difficult challenge due to the large solution space and unpredictability of opponents.
Classical and machine learning approaches should be investigated for their ability to overtake and avoid other cars.
Additionally, benchmark methods and standardised evaluations should be constructed to advance progress.

\bibliographystyle{IEEEtran}
\bibliography{IEEEabrv,ref}



\nopagebreak
\begin{appendices}

\section{Single-Track Model}\label{app:dynamic_model}
The F1TENTH-gym's simulation is based on a single-track model derived from the CommonRoad framework \cite{althoff2017commonroad}.
This model simplifies the vehicle kinematics to a two-wheel system but allows for under- and oversteering by introducing a slip angle $\beta$ and a friction coefficient $\mu$. It accounts for load transfer between axles, assumes no constant velocity, and is suitable for simulating manoeuvres close to the vehicle's physical driving limit. 

For a global position $[x, y]$ with yaw angle $\theta$, speed $v$ and steering angle $\delta$, the state is defined as
\begin{equation}
    \small
    \Vec{x} = 
    \begin{bmatrix}
        x & y & \delta & v & \theta & \dot{\theta} & \beta
    \end{bmatrix}^\top.
\end{equation}

The inputs into the system $u_1$ and $u_2$ are the steering angle velocity $v_{\delta}$, and longitudinal acceleration $a$.
Using the parameters of vehicle mass $m$, cornering stiffness $C_s$ and wheelbase length $l$ for the front $_f$ and rear $_r$, the height of the centre of gravity $h_\textbf{cg}$, and gravitational constant $g$, the state derivatives are given by,
\begin{equation}
\small
\begin{aligned}
\dot{x}_1 & = x_4 \cos(x_5+x_7), \quad\quad \dot{x}_2 = x_4 \sin(x_5+x_7), \\
\dot{x}_3 & = u_1, \quad\quad \dot{x}_4 = u_2, \quad\quad \dot{x}_5 = x_6\\
\dot{x}_6 & = \frac{\mu m}{I_z(l_r+l_f)}\big[l_f C_{S, f}(g l_r-u_2 h_{cg}) \cdot x_3 \\
           & \quad + (l_r C_{S, r}(g l_f+u_2 h_{cg}) - l_f C_{S, f}(g l_r-u_2 h_{cg})) \cdot x_7 \\
           & \quad - (l_f^2 C_{S, f}(g l_r-u_2 h_{cg})+l_r^2 C_{S, r}(g l_f+u_2 h_{cg})) \cdot \frac{x_6}{x_4}\big], \\
\dot{x}_7 & = \frac{\mu}{x_4(l_r+l_f)}\big[C_{S, f}(g l_r-u_2 h_{cg}) \cdot x_3 \\
           & \quad - (C_{S, r}(g l_f+u_2 h_{cg})+C_{S, f}(g l_r-u_2 h_{cg})) \cdot x_7 \\
           & \quad + (C_{S, r}(g l_f+u_2 h_{cg}) l_r-C_{S, f}(g l_r-u_2 h_{cg}) l_f) \cdot \frac{x_6}{x_4}\big] \\
           & \quad -x_6.
\end{aligned}
\end{equation}
Due to singularities, a kinematic model is used for $|v| < 0.1$.

\section{Map Characteristics} \label{app:map_information}

We quantify the characteristics of these maps by providing the length, straight percentage, and number of corners in Table \ref{tab:map_data}.
The straight percentage is the length of the track with curvature below 0.1~rad/m divided by the total track length.
The number of corners is the number of sections with curvature above 0.6~rad/m.
The table shows that the ESP is the longest track with a length of 236.93~m, AUT has the highest \% of straight sections and MCO has 16 corners, more than double the other maps.

\begin{table}[h]
    \centering
    \renewcommand{\arraystretch}{1.2}
    \begin{tabular}{w{l}{2cm} w{c}{1.1cm} w{c}{1.1cm} w{c}{1.1cm} w{c}{1.1cm}}
        \toprule
        Map name & AUT & ESP & GBR & MCO \\
        \midrule
        Track length [m] & 94.90 & 236.93 & 201.84 & 178.71 \\
        Straight length [\%] & 64.92 & 58.98 & 59.45 & 60.60 \\
        Corner count & 7 & 7 & 7 & 16 \\
        \bottomrule
    \end{tabular}
    \vspace{3mm}
    \caption{Quantifiable track characteristics}
    \label{tab:map_data}
\end{table}

\section{Particle Filter Tuning} \label{app:particle_filter_tuning}

\begin{figure}[h]
    \centering
    \includegraphics[width=\linewidth]{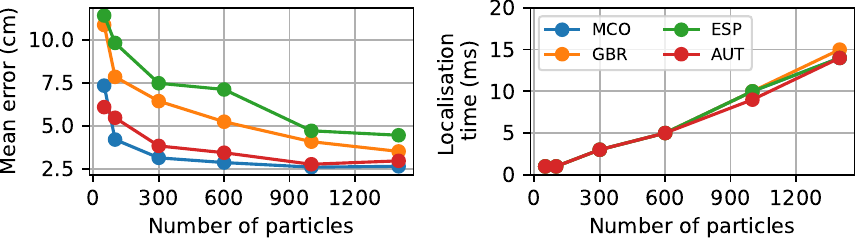}
    \caption{Localisation error and computation time w.r.t the number of particles.}
    \label{fig:pf_error_vs_particles}
\end{figure}

We ran a test lap using the pure pursuit algorithm following the centre line at a low constant speed of 2~m/s, and evaluated the accuracy using increasing numbers of particles.
We present Fig. \ref{fig:pf_error_vs_particles} to indicate the impact of the number of particles on the localisation accuracy and computation time.
The graph shows that the error drops from 10~cm for 50 particles to around 4~cm for 1000 particles. 
Using 1000 particles has a computation time of 10~ms, which is below our planning period of 40~ms.
Therefore, we use 1000 particles for our tests.

\end{appendices}

\newpage

\begin{IEEEbiography}[{\includegraphics[width=1in]{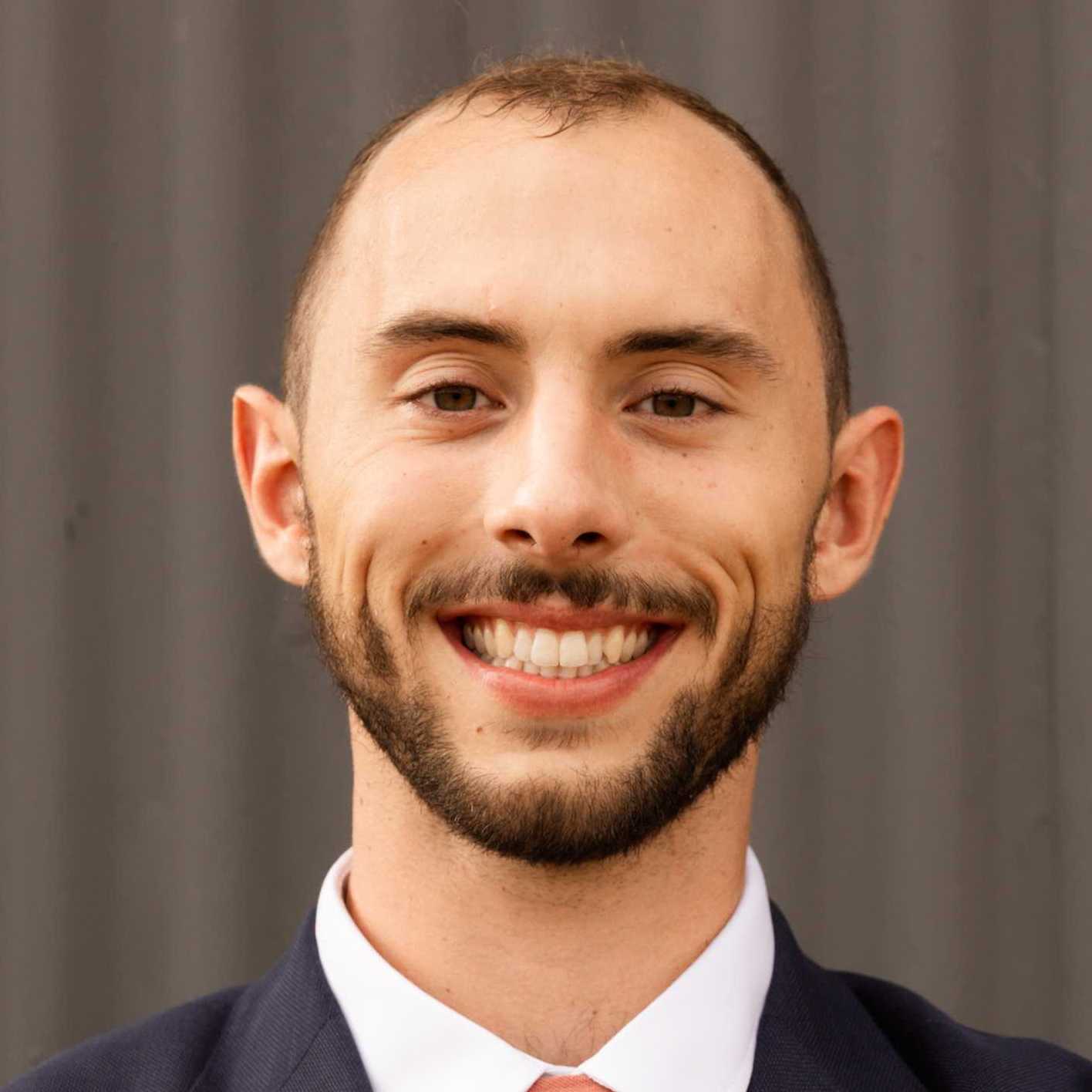}}]{Benjamin David Evans}
    studied for a bachelor's in Mechatronic Engineering at the University of Stellenbosch, graduating in 2019. 
    He started a masters, focused on using reinforcement learning for autonomous racing, which was upgraded to a PhD.
    After graduating in 2023, he continued as a postdoctoral researcher at Stellenbosch University.
    His interests include the intersection of classical control and machine learning for autonomous systems.
\end{IEEEbiography}

\begin{IEEEbiography}[{\includegraphics[width=1in]{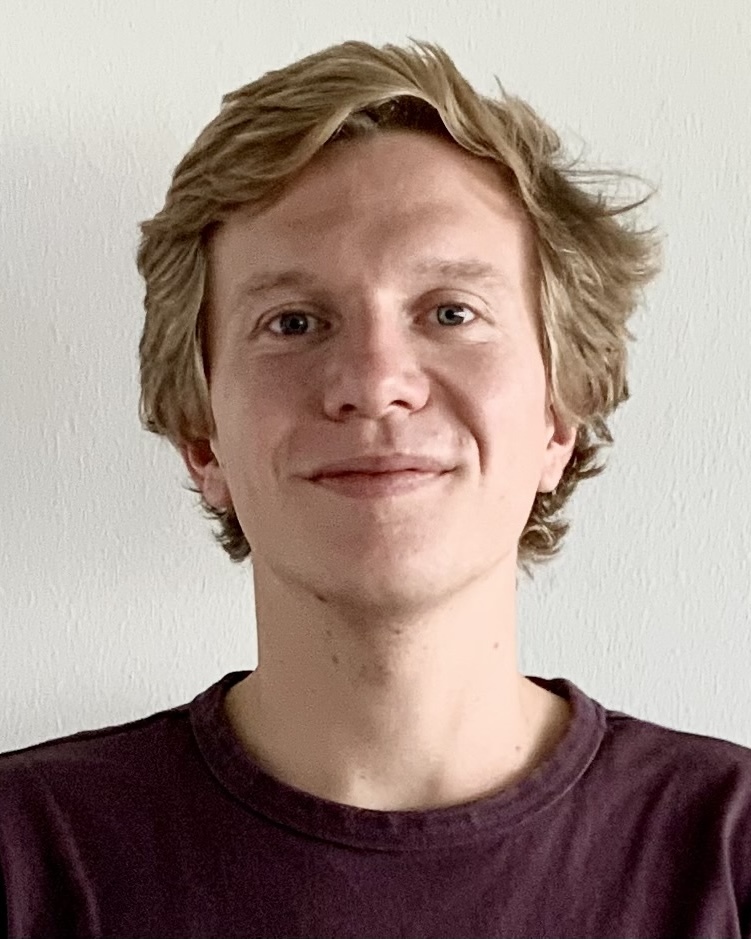}}]{Raphael Trumpp}
    graduated with a M.Sc. degree in mechanical engineering from the Technical University of Munich (Germany) in 2021, where he is currently pursuing a Ph.D. in informatics. His research focuses on machine learning, especially combining deep reinforcement learning with classical control methods for interactive multi-agent scenarios.
\end{IEEEbiography}

\begin{IEEEbiography}[{\includegraphics[width=1in]{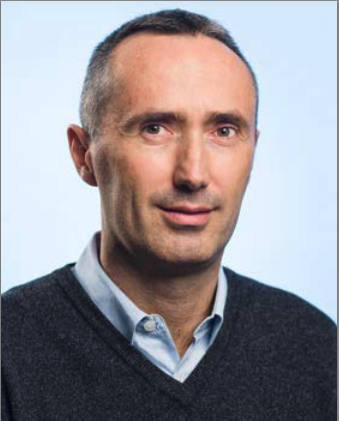}}]{Marco Caccamo}
earned his Ph.D. in computer engineering from Scuola Superiore Sant’Anna (Italy) in 2002. Shortly after graduation, he joined University of Illinois at Urbana-Champaign as assistant professor in Computer Science and was promoted to full professor in 2014. Since 2018, Prof. Caccamo has been appointed to the chair of Cyber-Physical Systems in Production Engineering at Technical University of Munich (Germany). In 2003, he was awarded an NSF CAREER Award. He is a recipient of the Alexander von Humboldt Professorship and he is IEEE Fellow.
\end{IEEEbiography}

\begin{IEEEbiography}[{\includegraphics[width=1in]{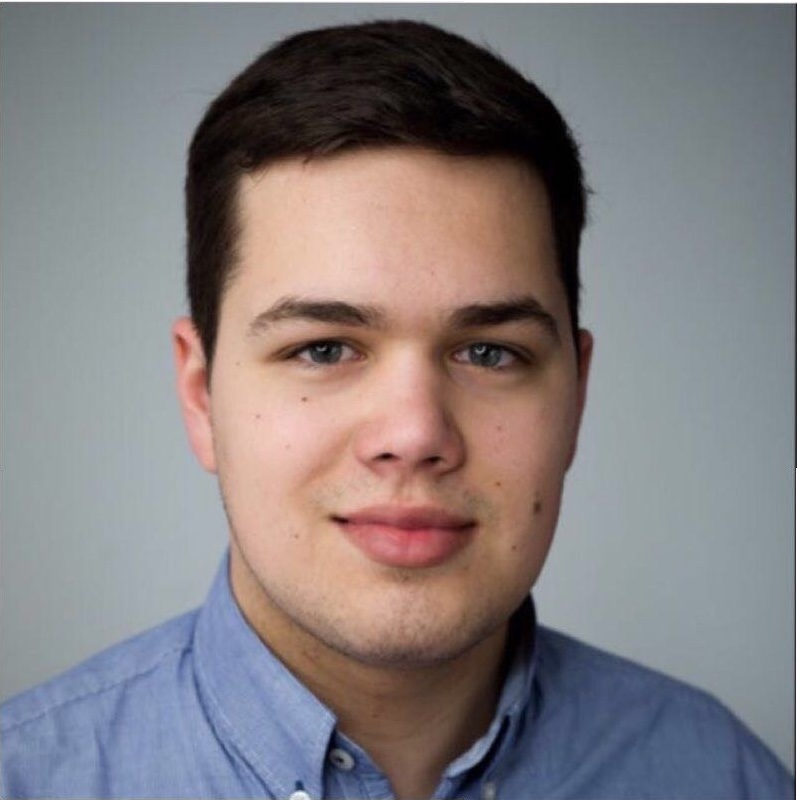}}]{Felix Jahncke}
received his bachelor's degree in Mechanical Engineering and master's degree in Automotive Engineering from the Technical University of Munich (TUM). During his master's studies, he completed research visits at the Technical University of Delft (Netherlands) and the University of Pennsylvania (USA). Since 2023, he has been pursuing his Ph.D. at TUM, focusing on the transition from traditional software architectures towards end-to-end learning methods, using F1TENTH vehicles.
\end{IEEEbiography}

\begin{IEEEbiography}[{\includegraphics[width=1in]{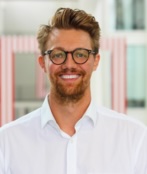}}]{Johannes Betz}
    is an assistant professor in the Department of Mobility Systems Engineering at the Technical University of Munich. He is one of the founders of the TUM Autonomous Motorsport team. His research focuses on developing adaptive dynamic path planning and control algorithms, decision-making algorithms that work under high uncertainty in multi-agent environments, and validating the algorithms on real-world robotic systems. Johannes earned a B.Eng. (2011) from the University of Applied Science Coburg, a M.Sc. (2012) from the University of Bayreuth, a M.A. (2021) in philosophy from TUM and a Ph.D. (2019) from TUM. 
\end{IEEEbiography}

\begin{IEEEbiography}[{\includegraphics[width=1in]{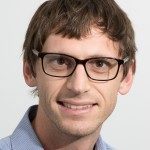}}]{Hendrick Willem Jordaan}
    received his bachelor's in Electrical and Electronic Engineering with Computer Science and continued to receive his Ph.D degree in satellite control at Stellenbosch University.  He currently acts as a senior lecturer at Stellenbosch University and is involved in several research projects regarding advanced control systems as applied to different autonomous vehicles.  His interests include robust and adaptive control systems applied to practical vehicles. He is a senior member of IEEE.
\end{IEEEbiography}

\begin{IEEEbiography}[{\includegraphics[width=1in]{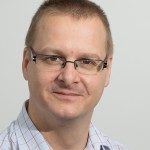}}]{Herman Arnold Engelbrecht}
   received his Ph.D. degree in Electronic Engineering from Stellenbosch University (South Africa) in 2007. He is currently the Chair of the Department of Electrical and Electronic Engineering. His research interests include distributed systems (specifically infrastructure to support massive multi-user virtual environments) and machine learning (specifically deep reinforcement learning). Prof Engelbrecht is a Senior Member of the IEEE and a Member of the ACM. 
\end{IEEEbiography}

\vfill

\end{document}